\pdfoutput=1

\documentclass[letterpaper, 10pt, journal]{ieeeconf}      %

\IEEEoverridecommandlockouts                              %

\let\labelindent\relax                                    %

\usepackage{graphicx}            %
\usepackage{amsmath}             %
\usepackage{amssymb}             %
\usepackage{amsfonts}            %
\usepackage{enumitem}            %
\usepackage{multirow}            %
\usepackage{siunitx}             %
\usepackage{url}                 %
\usepackage{xspace}              %
\usepackage[T1]{fontenc}         %
\usepackage[hidelinks]{hyperref} %
\usepackage{balance}
\usepackage[bottom]{footmisc}

\graphicspath{{images/}}
\DeclareGraphicsExtensions{.pdf,.png,.jpg,.jpeg}

\newcommand{\degreem}{^{\circ}} %

\newcommand{\figlabel}[1]{\label{fig:#1}}
\newcommand{\tablabel}[1]{\label{tab:#1}}

\newcommand{\figref}[1]{Fig.~\ref{fig:#1}\xspace}
\newcommand{\tabref}[1]{Table~\ref{tab:#1}\xspace}

\newcommand{\nop}{NimbRo\protect\nobreakdash-OP\xspace}
\newcommand{\nopxl}{NimbRo\protect\nobreakdash-OP2\xspace}
\newcommand{\dop}{DARwIn\protect\nobreakdash-OP\xspace}
\newcommand{\rop}{ROBOTIS OP2\xspace}
\newcommand{\nao}{Nao\xspace}
\newcommand{\cm}{CM730\xspace}
\newcommand{\cmnew}{CM740\xspace}

\newcommand{\igus}{igus\textsuperscript{\tiny\circledR}\xspace}
\newcommand{\iguhop}{igus\textsuperscript{\tiny\circledR}$\!$ Humanoid Open Platform\xspace}

\newcommand{\term}[1]{\emph{#1}\xspace}
\newcommand{\degree}{$\degreem$\xspace}

\title{\LARGE \bf NimbRo-OP2:\\Grown-up 3D Printed Open Humanoid Platform for Research}

\author{Grzegorz Ficht, Philipp Allgeuer, Hafez Farazi and Sven Behnke%
\thanks{All authors are with the Autonomous Intelligent Systems (AIS) Group, Computer Science Institute VI,
        University of Bonn, Germany. Email: {\tt\small ficht@ais.uni-bonn.de}. This work was partially
        funded by grant BE 2556/13 of the German Research Foundation (DFG).}}

\usepackage{eso-pic}

\AtBeginDocument{\AddToShipoutPictureFG*{\AtTextUpperLeft{\put(0,\LenToUnit{9pt}){\parbox{\textwidth}{\centering\bfseries
International Conference on Humanoid Robots (Humanoids), Birmingham, England, 2017
}}}}}
\begin{document}

\bstctlcite{IEEEexample:BSTcontrol}

\maketitle
\thispagestyle{empty}
\pagestyle{empty}

\addtolength{\textheight}{-0.1in}
\begin{abstract}
The versatility of humanoid robots in locomotion, full-body motion, interaction with unmodified human environments, and intuitive human-robot interaction led to increased research interest. Multiple smaller platforms are available for research, but these require a miniaturized environment to interact with---and often the small scale of the robot diminishes the influence of factors which would have affected larger robots. 
Unfortunately, many research platforms in the larger size range are less affordable, more difficult to operate, maintain and modify, and very often closed-source. In this work, we introduce \nopxl, an affordable, fully open-source platform in terms of both hardware and software. 
Being almost 135\,cm tall and only 18\,kg in weight, the robot is not only capable of interacting in an environment meant for humans, but also easy and safe to operate and does not require a gantry when doing so.
The exoskeleton of the robot is 3D printed, which produces a lightweight and visually appealing design. 
We present all mechanical and electrical aspects of the robot, as well as some of the software features of our well-established open-source ROS software. The \nopxl performed at RoboCup 2017 in Nagoya, Japan, where it won the Humanoid League AdultSize Soccer competition and Technical Challenge.
\end{abstract}

\section{Introduction}

The idea that robots with a humanoid body plan, equipped with human-like motion and multimodal communication skills could relive humans from performing a large variety of tasks has been a key factor motivating the research on humanoid robots. Even though multiple platforms exist that allow researchers to explore this field, none of the available ones are affordable and large enough to allow for easy interaction with an unmodified human environment. To develop such a platform requires not only a lot of resources but also expertise, which is a steep entry point for any research team. As demonstrated by successful examples, such as Open-HRP~\cite{Kaneko2009}, Aldebaran Nao, and our own predecessor NimbRo-OP~\cite{allgeuer2015child}, having standard platforms openly available facilitates collaboration and accelerates research. 

\begin{figure}[!t]
\centering
\raisebox{3ex}{\includegraphics[width=0.52\linewidth]{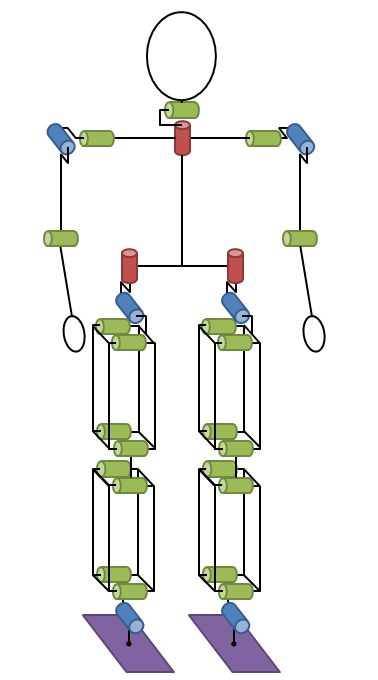}}\includegraphics[width=0.44\linewidth]{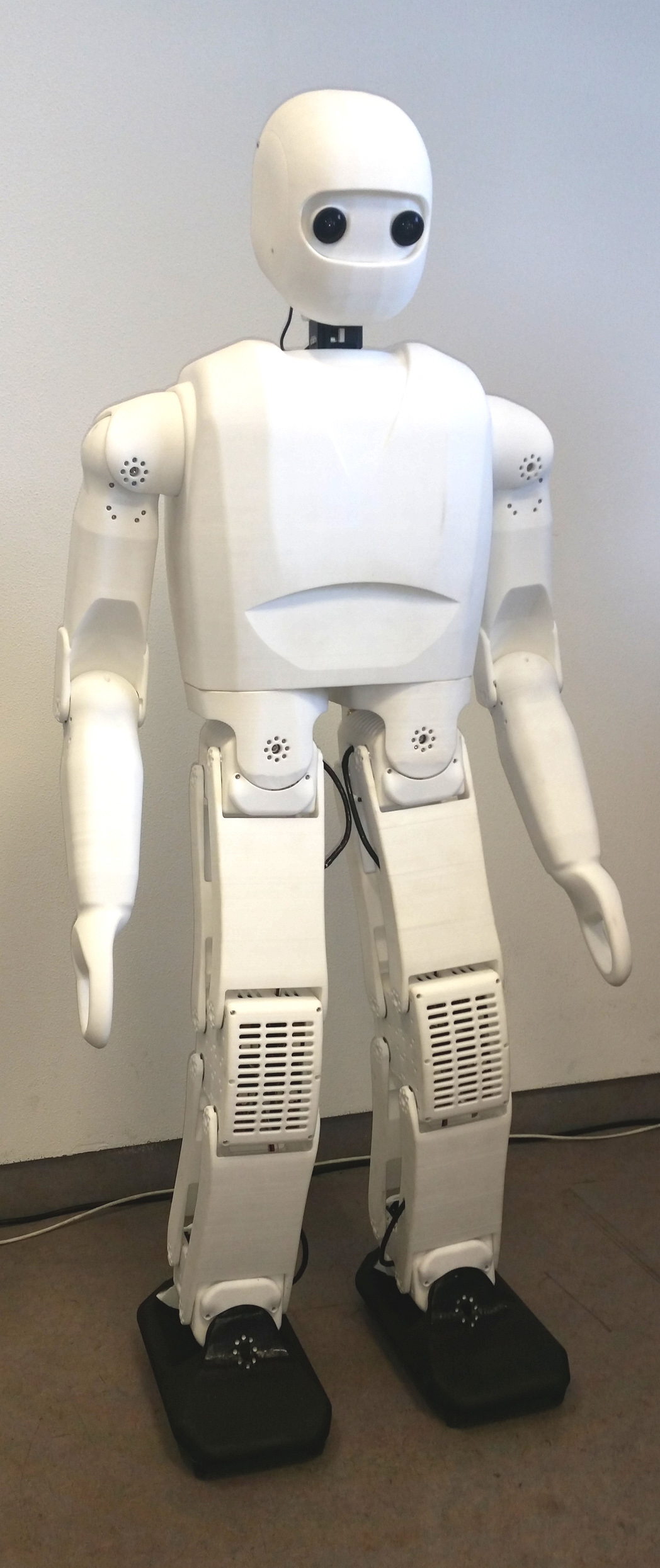}
\caption{The \nopxl: Kinematics and assembled robot.}
\label{OP2_teaser}
\end{figure}

The \nopxl, which we are presenting in this paper, is the effect of our many years of experience with building such platforms, like the \nop and the collaborative work on the \iguhop together with \igus GmbH. The \nopxl aims to fill the void for affordable, adult-sized humanoid robots without compromising on factors such as output torque or computing power.
Due to the 3D printed nature of the robot, parts can be modified and replaced with great freedom. The robot, along with its kinematics schema is shown in Fig.~\ref{OP2_teaser}. 
A demonstration video of the \nopxl is available\footnote{Video: 
\url{http://www.ais.uni-bonn.de/videos/Humanoids_2017}}.

\section{Related Work}
Humanoid robot design has been investigated by multiple industrial and academic research groups.
Several standard humanoid platforms for research are available. By using a platform which already handles low level tasks, such as joint actuation, and sensor feedback, researchers can directly focus on their specific research topics.
This can be most clearly observed for robots that are less than \SI{60}{cm} tall, as the cost of their acquisition is lowest. Examples include the \nao robot~\cite{Gouaillier2009}, from Aldebaran Robotics and both, the \dop \cite{Ha2011} and its successor the \rop, distributed by Robotis.
These robots have enabled many groups to use them as the foundation for their research, thanks to the available software and documentation. Being small gives the robot an advantage of also being low-cost. Small size is also a limiting factor when it comes to application, however, as small robots require a scaled-down environment.
Recently, also slightly taller humanoid platforms have become available. These platforms could be considered to be child-sized so their interaction with a human environment is possible---although limited. The two most prominent child-size examples are the \SI{84}{cm} tall Poppy robot from the Inria Flowers Laboratory~\cite{Lapeyre2014} and the result of our collaboration with \igus GmbH: the \SI{90}{cm} tall \iguhop~\cite{allgeuer2015child}. Both of these robots are open source platforms that are mostly 3D printed. Poppy's main features include a compliant bio-inspired morphology, a multi-articulated trunk and skeletonized design.
Although visually appealing, Poppy has not yet displayed walking without assistance, nor with on-board power supply and computing. In our previous work, we have shown the \iguhop performing various tasks, such as autonomous omni-directional balanced walking, or getting up from different poses, without external help.

When comparing the previously mentioned robots to larger humanoid platforms, such as Honda Asimo~\cite{Hirai1998,hirose2007honda}, HRP~\cite{Kaneko2009}, Hubo~\cite{park2005mechanical}, NASA Valkyrie~\cite{radford2015valkyrie}, DLR TORO~\cite{englsberger2014overview}, and Boston Dynamics Petman~\cite{nelson2012petman} and Atlas robots, one can easily notice the increased complexity, which contributes to a price being two orders of magnitude higher.
By being more complex and expensive, these platforms are much more difficult to operate and maintain.
In extreme cases, they can even be dangerous to the user. These factors limit the possibility of using such robots by most research groups.

\begin{figure}[!t]
\centering\includegraphics[width=1.0\linewidth]{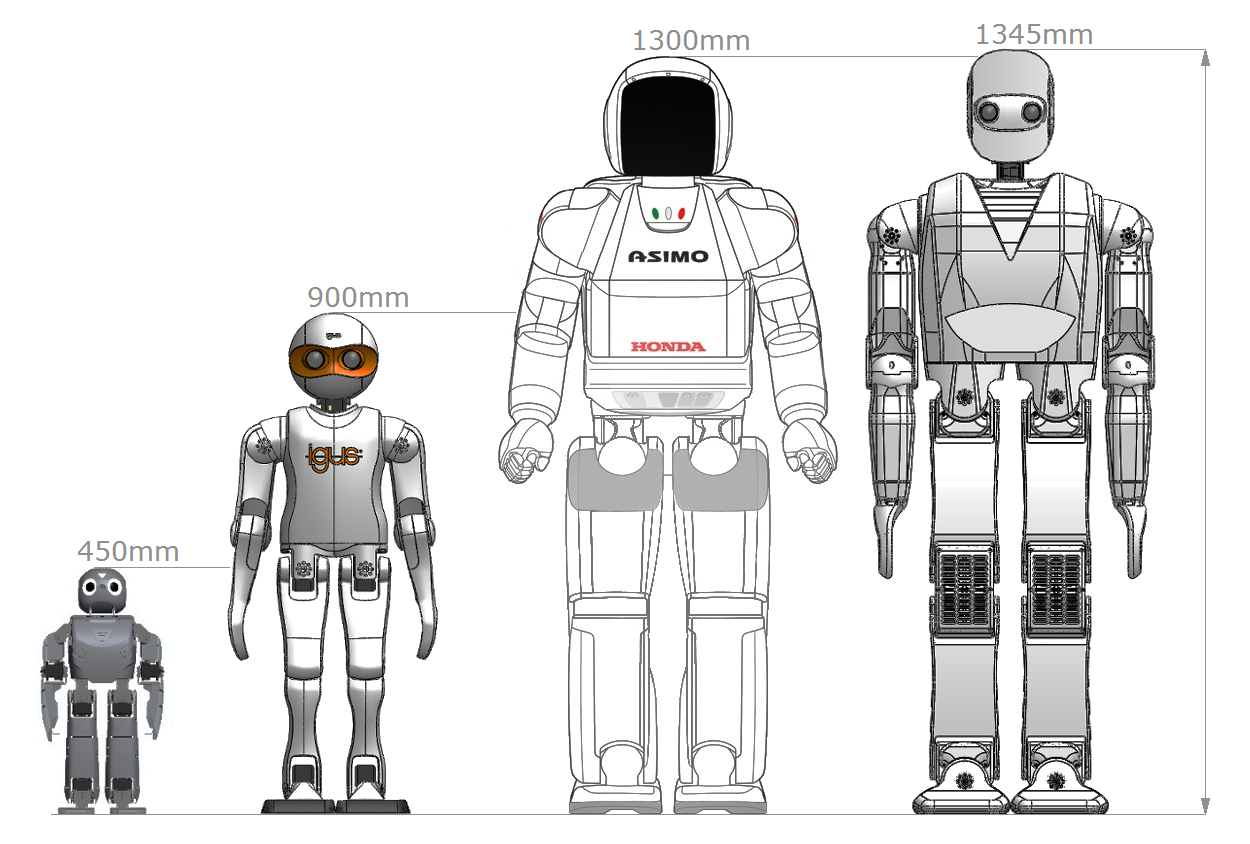}\vspace*{-1ex}
\caption{Approximate size comparison of various robot platforms, from left: \dop, \iguhop, Honda Asimo and the presented \nopxl.}
\label{OP2_comparison}
\end{figure}

The \nopxl addresses the issues mentioned with the robots above and seeks to be the solution for an affordable, maintainable and customizable larger-sized, humanoid robot. 
A comparison between different sized platforms and the \nopxl is shown in Fig.~\ref{OP2_comparison}.
The cost of the NimbRo-OP2 is about twice of the 90\,cm \iguhop~\cite{allgeuer2015child} and about four times that of the 45\,cm \dop, but at least one order of magnitude less than humanoid robots of similar size. 

\section{Hardware Design}

\begin{table}
\renewcommand{\arraystretch}{1.2}
\caption{\nopxl specifications.}
\tablabel{OP2_specs}
\centering
\footnotesize
\begin{tabular}{c c c}
\hline
\textbf{Type} & \textbf{Specification} & \textbf{Value}\\
\hline
\multirow{4}{*}{\textbf{General}} & Height \& Weight & \SI{1345}{mm}, \SI{17.5}{kg}\\
& Battery & 4-cell LiPo (\SI{14.8}{V}, \SI{6.6}{Ah})\\
& Battery life & \SI{15}{}--\SI{30}{\minute}\\
& Material & Polyamide 12 (PA12)\\
\hline
\multirow{5}{*}{\textbf{PC}} & Product & Intel NUC NUC7I7BNH\\
& CPU & Intel Core i7-7567U, \SI{3.5}{}--\SI{4.0}{GHz}\\
& Memory & \SI{4}{GB} DDR3 RAM, \SI{120}{GB} SSD\\
& Network & Ethernet, Wi-Fi, Bluetooth\\
& Other & 4$\,\times\,$USB 3.0, HDMI, Thunderbolt 3\\
\hline
\multirow{3}{*}{\textbf{\cm}} & Microcontroller & STM32F103RE (Cortex M3)\\
& Memory & \SI{512}{KB} Flash, \SI{64}{KB} SRAM\\
& Other & 3$\,\times\,$Buttons, 7$\,\times\,$Status LEDs\\
\hline
\multirow{6}{*}{\textbf{Actuators}} 
& Stall torque & \SI{10.0}{Nm} \\
& No load speed & \SI{55}{rpm} \\
& Total & 34$\,\times\,$MX-106R\\
& Neck & 2$\,\times\,$MX-106R\\
& Each arm & 3$\,\times\,$MX-106R\\
& Each leg & 13$\,\times\,$MX-106R\\
\hline
\multirow{5}{*}{\textbf{Sensors}} & Encoders & \SI{4096}{ticks/rev}\\
& Gyroscope & 3-axis (L3G4200D chip)\\
& Accelerometer & 3-axis (LIS331DLH chip)\\
& Camera & Logitech C905 (720p)\\
& Camera lens & Wide-angle lens with 150\degree\!FOV\\
\hline
\end{tabular}
\end{table}
As with our previous platforms, we have focused on creating a design that is easy to manufacture, assemble, maintain, and modify. The usage of 3D-printing technology contributes a great amount to all of these points. For a summary of the main hardware components, please refer to \tabref{OP2_specs}. 

\subsection{Mechanical Structure} %

To create an affordable platform of such size which has enough torque not only to support its own weight but is also able to walk and perform dynamic motions we have decided to use a kinematic structure which efficiently combines multiple actuators working in parallel, external gearing and regular serial chains. 
By using only one COTS (commercial off-the-shelf) type of actuator we have not only minimized complexity, but reduced cost as well, as the actuators can be purchased in bulk.

The frame of the new design is completely 3D-printed with Selective Laser Sintering (SLS) performed on Polyamide 12 in approximately \SI{0.1}{mm} increments. The robot's structure is supported without any additional frames behind the outer surface---the exoskeleton is not only load-bearing but also for outward appearance. This approach greatly increases space and weight savings, as evidenced by the very low weight of the \nopxl relative to its size. The structural integrity and resistance to deformation and buckling is ensured through varying the wall thickness in the areas that need it, and through regular widespread use of ribs and other geometric strengthening features, which are printed directly into the structure. Essentially, through the freedoms of 3D printing, the durability of the frame can be maximized exactly where it is needed, without making the whole structure overly heavy. The CAD files are available online~\cite{NOP2Hardware}. 

\begin{figure}[tbh]
\centering\includegraphics[width=0.51\linewidth]{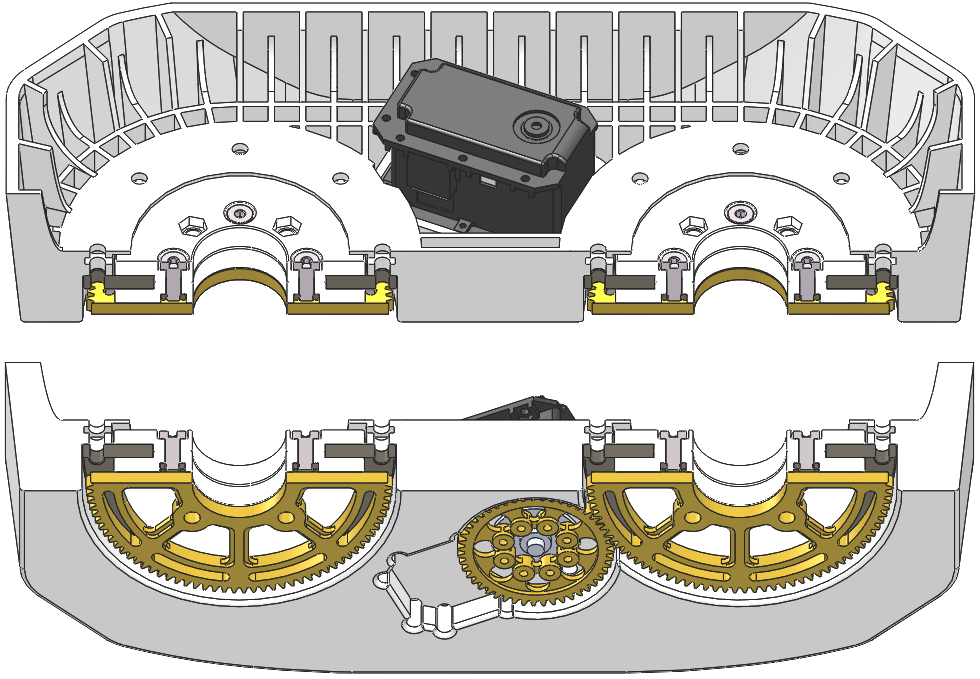}\hspace*{2ex}\includegraphics[width=0.46\linewidth]{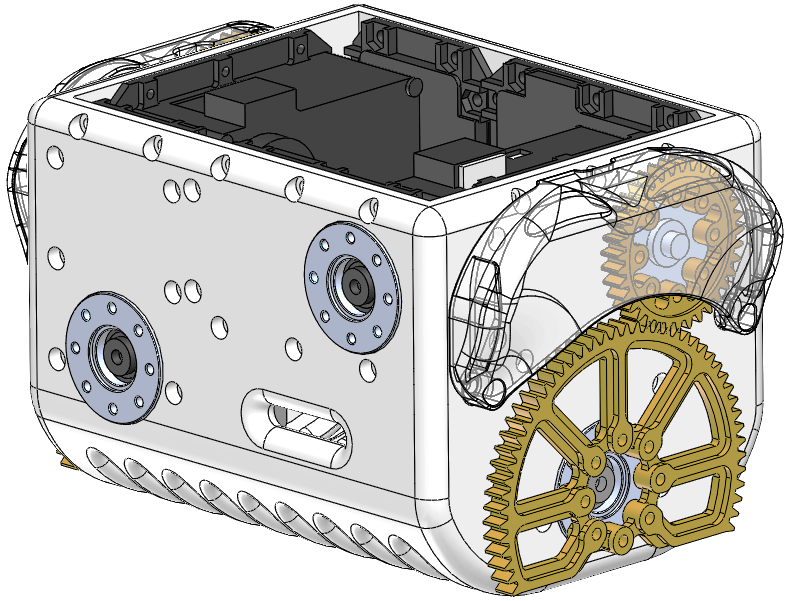}
\caption{Joints using external gearing. Left: Hip cross-section from top and bottom. Right: Hip and ankle Joint (gear cover shown is transparent).}
\label{OP2_gearjoints}
\end{figure}

\begin{figure}[tbh]

\centering
\includegraphics[height=0.6\linewidth]{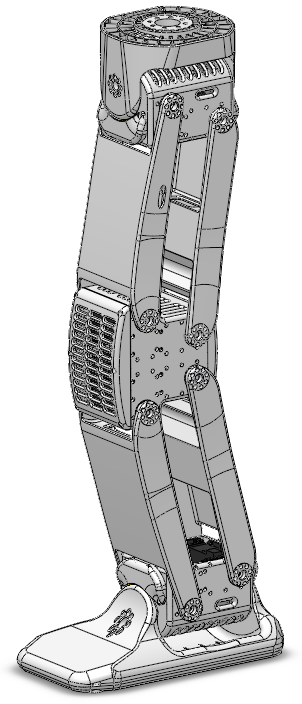}\hspace*{1ex}
\includegraphics[height=0.35\linewidth]{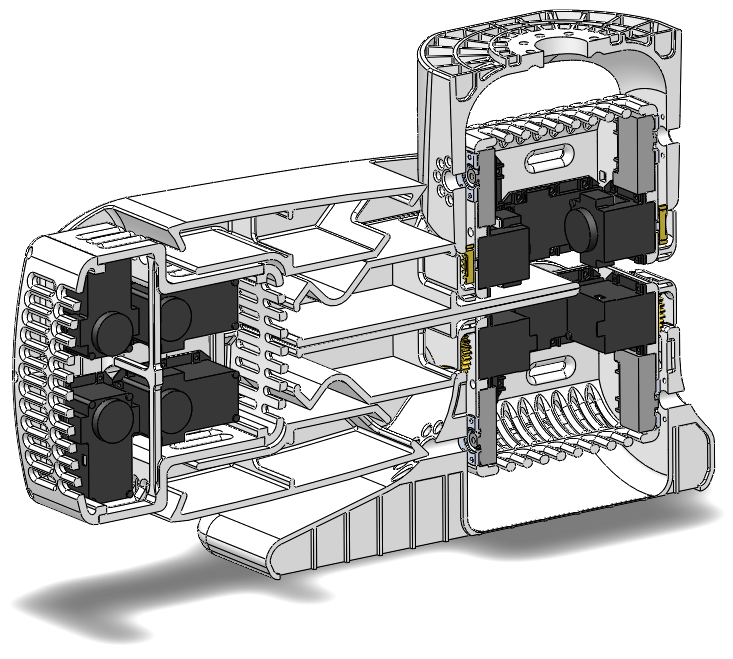}\hspace*{1ex}
\includegraphics[height=0.32\linewidth]{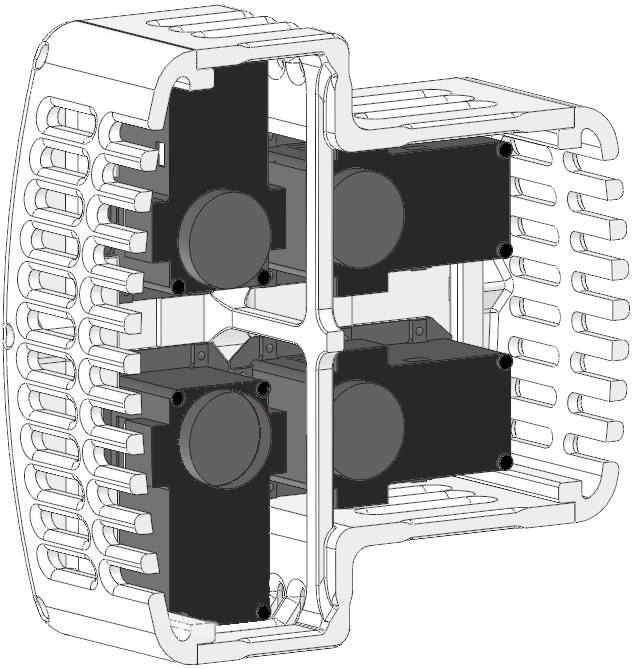}\vspace*{-1ex}
\caption{Parallel kinematics leg design. Left: Fully extended; Middle: Fully shortened; Right: Cross-section of the knee.}
\label{OP2_legdesign}
\end{figure}

\subsection{Design Optimizations} %

The design of the robot has been optimized in many ways, which contribute to the overall performance and appearance. All actuators are easily accessible for diagnostics and repairs either directly or after a cover removal. All of the leg servos---which are the most probable to be under stress in normal operation---have the possibility to dissipate heat through strategic placement of venting holes. The knee is the best example of both of these points. A single knee is composed out of eight actuators and three plastic parts. 
Four of the servos are accessible from the front and four from the back. All three 3D-printed components have vents in the most crucial points, while still remaining rigid enough to support the torque being exerted. For extra cooling, mounting DC fans is also possible, but so far was not necessary. 
Having universal parts was also a point of the design, so that a single part could act as a spare in multiple spots. The parallel kinematics in the thigh and shank are a representation of this. The parts are horizontal and vertical mirror images of each other, so although the thigh and shank consist of eight parts in total, having two is enough to repair the legs in case any of the parts should break. The same can be said about the upper and lower arm and their covers, which are the same in both the shoulder and elbow joints.

\subsection{Actuation Methods} %

The kinematic structure of the robot is shown in Fig.~\ref{OP2_teaser}. The upper body only consists of serial chains. The neck is formed by a yaw and a pitch joint, while the arms have two pitch joints with a roll joint in between. To reinforce the arms, two \igus axial thrust bearings have been used in between the trunk and shoulder. This solution not only provides the required dry-rubbing self-lubrication but also removes any possible stress from the shoulder pitch axis, increasing the lifespan of the actuators. 
The noteworthy differences that separate the \nopxl and the \iguhop are in the construction of the legs.
Although we use only one type of actuator, through their arrangement in combination with gearing we can distinguish three different joint types. The first type has been employed in the hip yaw and consists of a single actuator and external gearing. This gear is mounted in between the hip part and an \igus PRT slewing ring bearing, which connects the hip to the torso. The three components all have a common hole going through the middle that allows for effortless cable feeding.
The second actuation type is used in both the hip and ankle roll, as these joints mirror each other. Here, two actuators work in a master-slave connection that drives a set of external gears. The final type is used in the parallel kinematics and consists of fitting an actuator into each mounting point of the thigh and shank and using a master-slave connection to maintain synchronization between the servos.
The various type of joints used and their design can be seen in Fig.~\ref{OP2_gearjoints} and Fig.~\ref{OP2_legdesign}.

\begin{figure}[!t]
\parbox{\linewidth}{\centering\includegraphics[height=0.65\linewidth,width=0.95\linewidth]{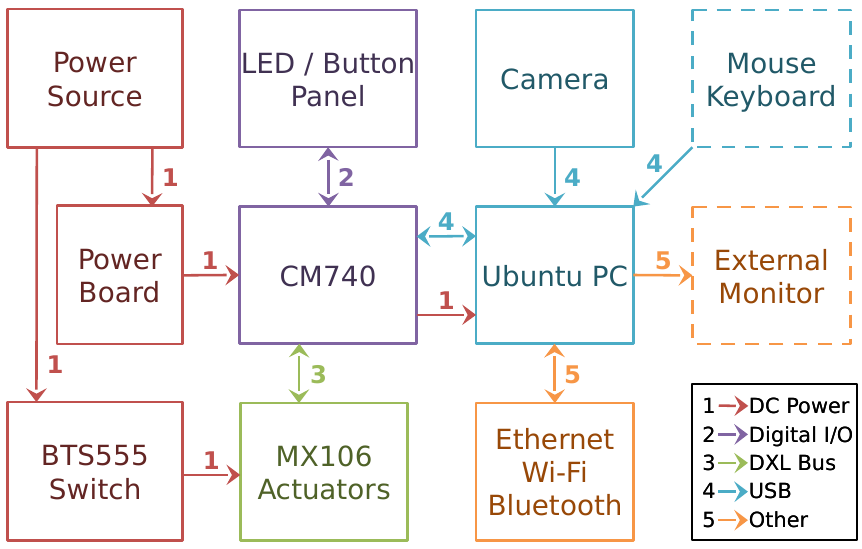}}
\caption{Electrical scheme of the \nopxl.}
\figlabel{electrical_architecture}
\end{figure}

\subsection{Robot Electronics}

The key to having a low-cost and capable platform is lowering the complexity without compromising on any features. The electronics of the robot are a perfect reflection of this mindset. At the core of the computing unit lies the newest Intel NUC computer with an Intel Core~i7-7567U processor operating at a possible maximum frequency of 4.0GHz.  The onboard PC is not limited to currently available small form factor (SFF) PCs, but in the design we have taken into account future possible upgrades by incorporating VESA75 and VESA100 standard mounts. 
In combination with a non-specific back cover (unlike in the \iguhop), the user is granted great flexibility when choosing a SFF PC. If even more compute power is needed, there is enough space inside the robot torso to fit a Mini-ITX board with a GPU. Detailed specifications are given in \tabref{OP2_specs}.

To control all of the Dynamixel MX-106R actuators on the RS-485 bus, we employ the \cmnew sub controller board provided by the same manufacturer---Robotis, from South Korea.
The sub controller consists of an STM32F103RE microcontroller with on-board peripherals, such as a 3-axis accelerometer, 3-axis gyroscope, RS-485 transceiver and USB driver. To increase reliability and performance, the firmware has been redesigned and rewritten when developing the \iguhop.
Since our first release, many new features have been added e.g. recovery after a power failure.
Compatibility with the standard Dynamixel protocol has been retained.

Interfacing with the robot is done with a sub-board that is included in the \cmnew set.
Currently, the three available buttons on the board are used to fade-in and fade-out the robot, start and stop the behaviour control and force a power reset to relax the servos in case of an emergency. These functions can be reprogrammed and additional functionality through pushing and holding the buttons can also be achieved. A total of seven LEDs provide the user with feedback on the state of the robot, two of which are RGB controlled.

Further available external connections to the robot include USB, HDMI, Mini DisplayPort, Gigabit Ethernet, IEEE 802.11b/g/n Wi-Fi, and Bluetooth 4.0. By default, the Ethernet network interface of the PC is configured system-wide to automatically swap between static and DHCP connections as required, while simultaneously allowing the use of a static secondary 
IP address while DHCP is active.
\begin{figure}[t]
\parbox{\linewidth}{\centering\includegraphics[height=0.78\linewidth,width=0.95\linewidth]{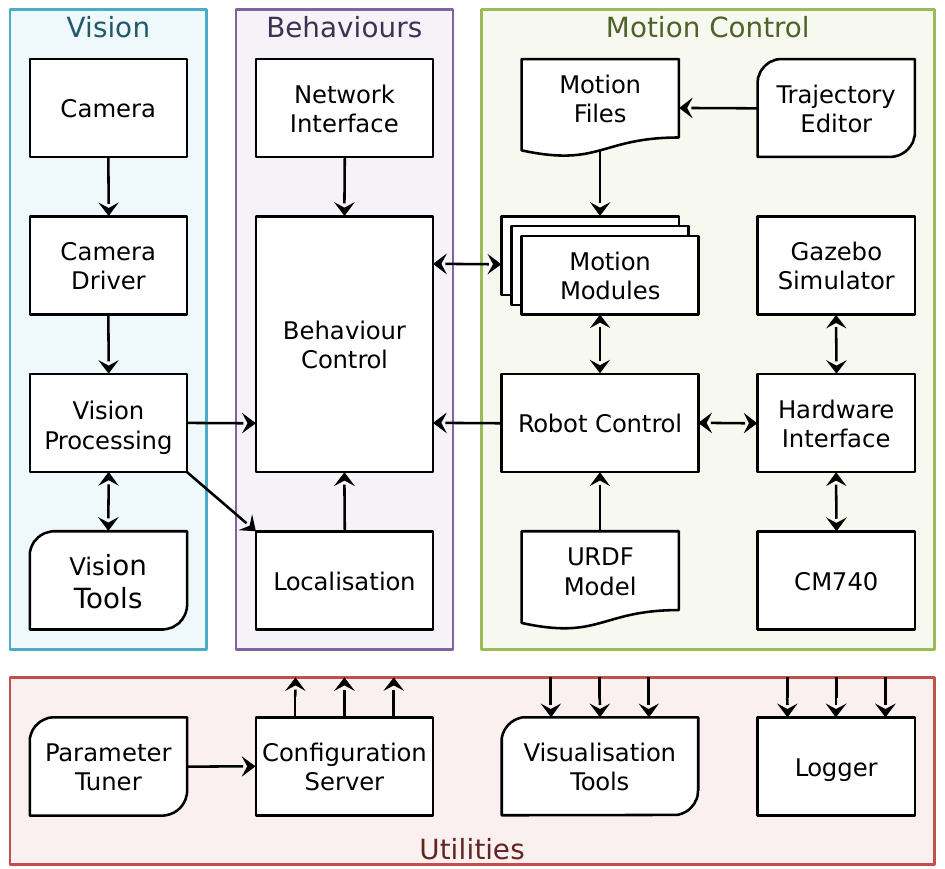}}
\caption{Architecture of our open-source ROS software.}
\figlabel{software_architecture}
\end{figure}
A very inconvenient design feature of the \iguhop was the power board which was originally intended to use with the \dop and \rop, which consume less current. The advantage of using it was the ability to apply power to the whole robot with a single switch. However, due to inrush currents there exists a potential risk of damaging the electronic components. This effect scales in size with the number of devices connected to the power supply. The power board also has a current limit which we exceed with the amount of actuators used. To solve this, we have created a power switch circuit based on the BTS555 smart high-side power switch which separates the actuators from the control components. 
This solution has the added benefit of being able to make working changes on the robot without unnecessarily dissipating power through idle servos. The Robotis power board is still used for the control components as it is a part of the microcontroller board set. The full electrical architecture can be seen in \figref{electrical_architecture}.

For vision purposes a single USB camera is used, which is mounted in the right eye. The vision can be expanded to stereo with a second camera in the left eye. For compatibility purposes between all of our platforms, we have decided to use the 720p Logitech C905 camera with a 150\degree wide-angle lens. However, the user has the flexibility to mount any other camera depending on the requirements, as some teams have done with the \iguhop, thanks to the flexible nature of 3D-printing.

\section{Software} %

Our open-source software based on the ROS middleware~\cite{Quigley2009} has become a well-established framework in the research community since the initial release, with multiple views and downloads each week. We continue to further develop it, in hope that other research groups can contribute to the goal of creating a user-friendly and effortless robot experience.

\subsection{Framework} %

The software architecture that we use on the \nopxl continues to expand on the legacy of the \nop and its successor, the \iguhop~\cite{Allgeuer2013a}. The sources are available online~\cite{IguhopSoftware}. Although the software was developed with humanoid robot soccer in mind~\cite{Schwarz2013}, the modularity of the behaviour control, motion modules and vision allows practically any other usage of the robot. For a full schematic of all the software interaction, please refer to \figref{software_architecture}. Many parts of the framework have been reused, as was the intention when developing the \iguhop---which proves the flexibility of our design. We build upon the inherent modularity of ROS with the usage of plugin schemes. By providing an interface to a selected task, the user has the liberty to implement his or her own solution or select from one of the provided plugins.

\subsection{Vision} %

\begin{figure}[t]
\parbox{\linewidth}{\centering
\includegraphics[width=0.50\linewidth,height=0.31\linewidth]{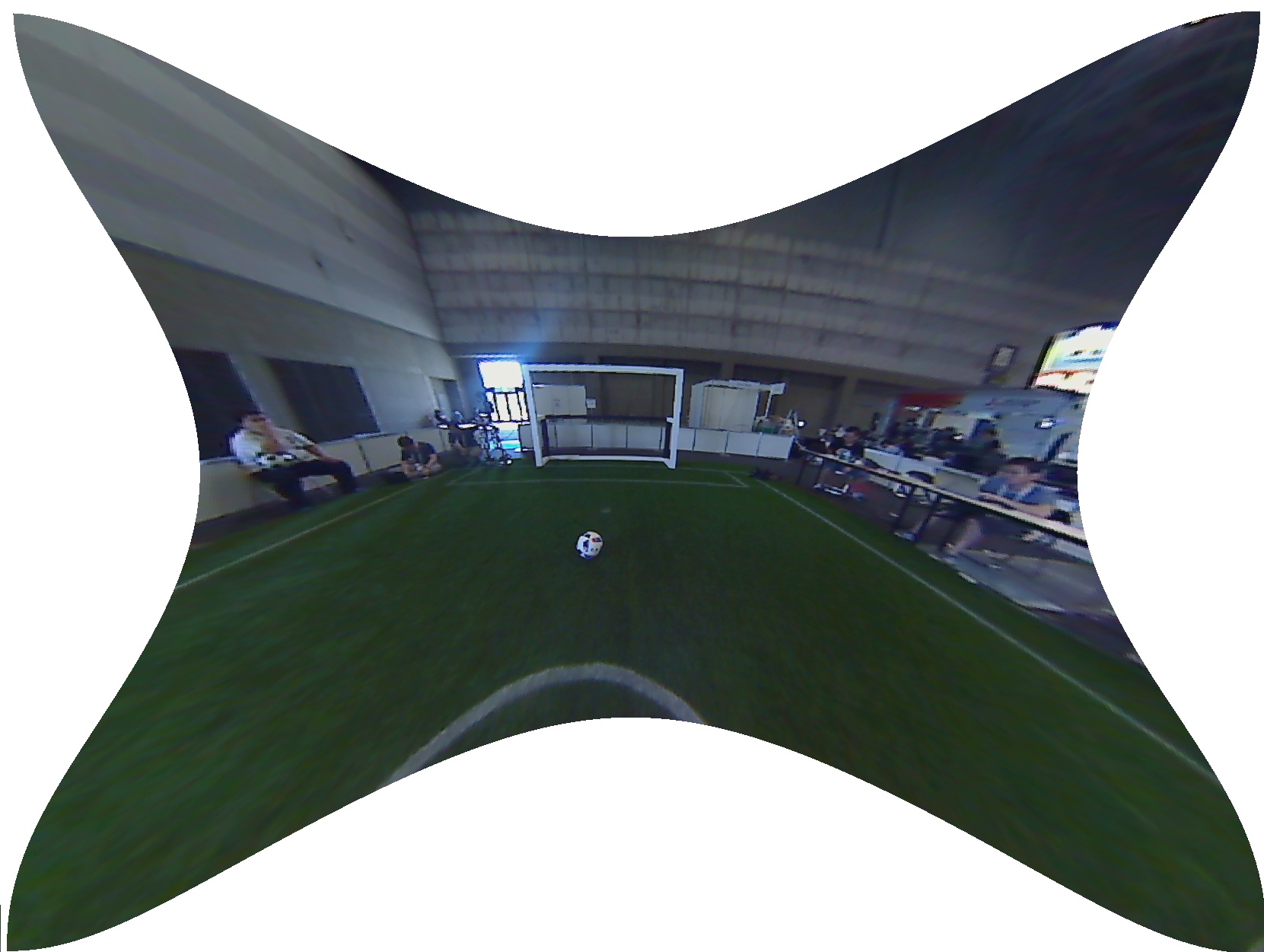}~\includegraphics[width=0.50\linewidth,height=0.32\linewidth]{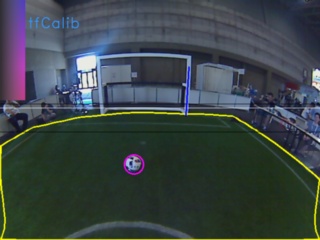}
}
\caption{Left: Undistorted image.
Right: Result of visual perception of soccer situation, including ball, field, and goal.}
\figlabel{vision_output}
\end{figure}

For visual perception of the environment, the selected camera captures the picture with a resolution of $1600\!\times\!1200$ at \SI{30}{Hz}. In order to achieve real-time processing the image is scaled down in ROS to a resolution of $640\!\times\!480$. We then convert it to the HSV color space to separate color data from luminance.
Due to the distortion introduced by the wide-angle lens, we use a Newton-Raphson based undistorting method which populates a pair of lookup tables, thus allowing real-time performance. The effect of undistorting the image is shown in \figref{vision_output}. With an undistorted image, we project the image coordinates into egocentric world coordinates. Although we have a relatively exact kinematic model of the robot, some unavoidable deviations still occur in the real hardware, potentially resulting in large projection errors for distant objects. 
To correct this, we calibrate the position and orientation of the camera frame in the robot's head using observations and the Nelder-Mead method. 

\subsection{Orientation Feedback}

To utilize closed-loop control, state feedback and estimation is necessary. In the \nopxl we use a 6-axis IMU built into the \cmnew sub controller board. We apply 3D nonlinear passive complementary filtering to raw measurements to achieve a globally stable 3D orientation output~\cite{Allgeuer2014}. We choose to represent this data as \term{fused angles}, as this representation is advantageous in the context of balance~\cite{Allgeuer2015}.

\begin{figure}[t]
\centering\includegraphics[width=1.0\linewidth]{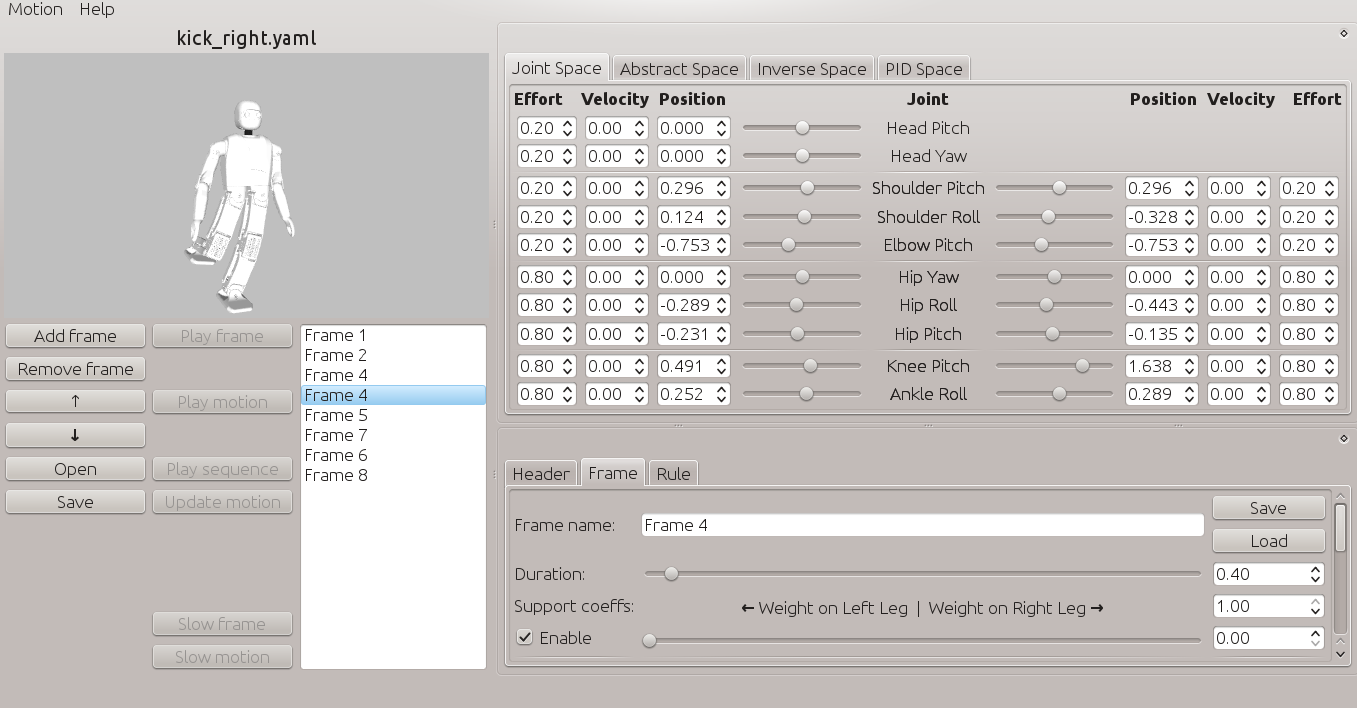}\vspace*{-1ex}
\caption{Trajectory editor used to develop motions for our humanoid robots, including the \nopxl.}
\label{trajectory_editor}
\end{figure}
\subsection{Model-based Feed-forward Position Control}

Reliably performing complex motions requires high tracking precision. To counter the effect of applying load to the joints, we use a feed-forward control scheme which modifies the joint trajectories based on the input trajectory and expected loads~\cite{Schwarz2013a}.
To calculate the feed-forward components to offset the trajectories, factors such as battery voltage, joint friction and inertia of the model are used. To obtain this data we employ the Rigid Body Dynamics Library to calculate the full-body inverse dynamics of the robot~\cite{Felis2015}.
Each servo in the robot is configured to use exclusively proportional control. To control compliance of the joints we interpolate the proportional gain with an effort value between 0 and 1---higher effort values result in increased joint stiffness, through greater P gain values.
To ensure proper servo operation, the proportional gain value is coerced at the lowest communication level.

\subsection{Gait Generation}

The gait currently in use is similar to that of the \iguhop. It is based on an open loop pattern generated from the desired step frequency~\cite{Behnke2006}.  
Stabilization elements of gaits based on the open loop core and orientation feedback have been incorporated to keep the robot balanced, the most current one using fused angle feedback directly to add corrective actions to joint trajectories~\cite{Missura2013a}~\cite{Allgeuer2015}. Although these gaits have been developed with robots without parallel kinematics in mind, the gait is also compatible with the parallel kinematics of the \nopxl, through various conversions described in the next subsection.

\subsection{Parallel Kinematics Adaptation}

To maintain compatibility with our other robots and enhance the possibility of using parallel kinematics on future platforms we have developed a new hardware interface which creates a set of dependencies between chains in the URDF model. The URDF model of the \nopxl has two instances of left and right legs.
A virtual serial chain and a real parallel one. All of the conversions are set up automatically, based on joint name aliases from the URDF model and consist of copying, negating, summing and subtracting commands. Scaling of values for external gearing is also applied on this level. 
The conversions done include not only set joint positions but torques produced in the joint. 
These conversions work in both directions: when sending commands and reading feedback.

\begin{figure}[t]
\parbox{\linewidth}{\centering
\includegraphics[height=0.295\linewidth]{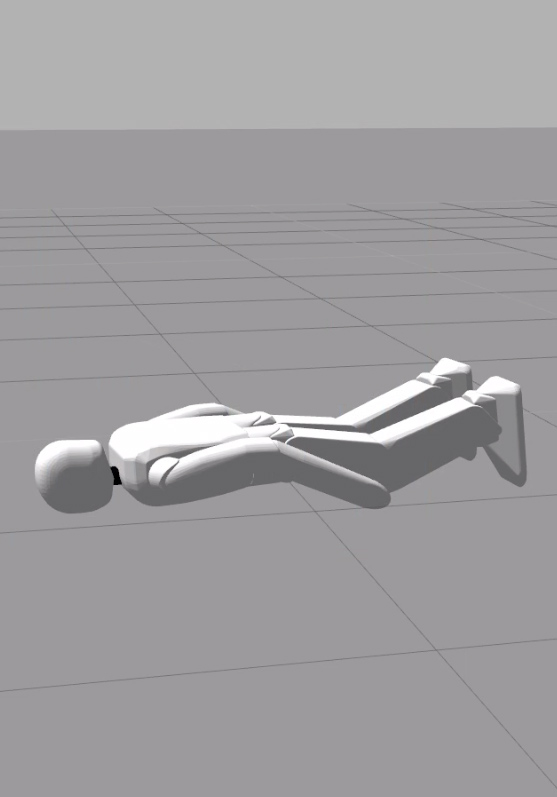}\hspace{0.5px}
\includegraphics[height=0.295\linewidth]{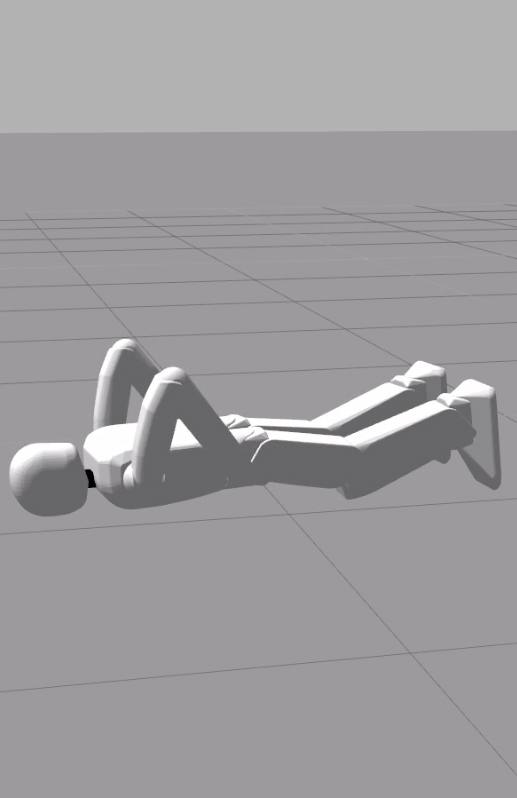}\hspace{0.5px}
\includegraphics[height=0.295\linewidth]{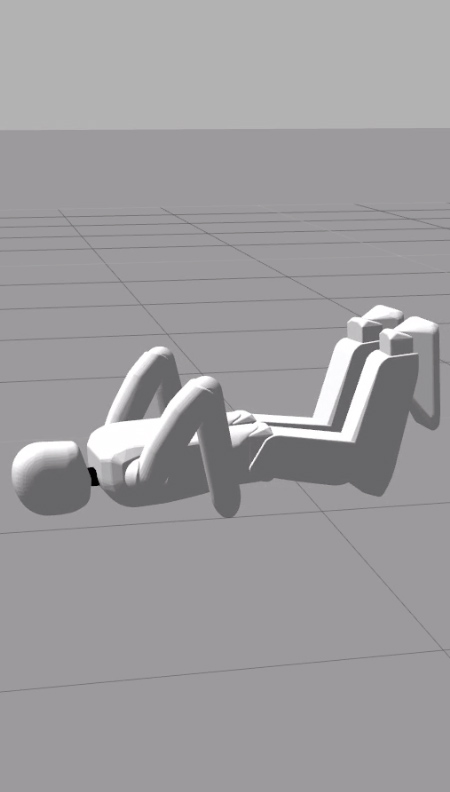}\hspace{0.5px}
\includegraphics[height=0.295\linewidth]{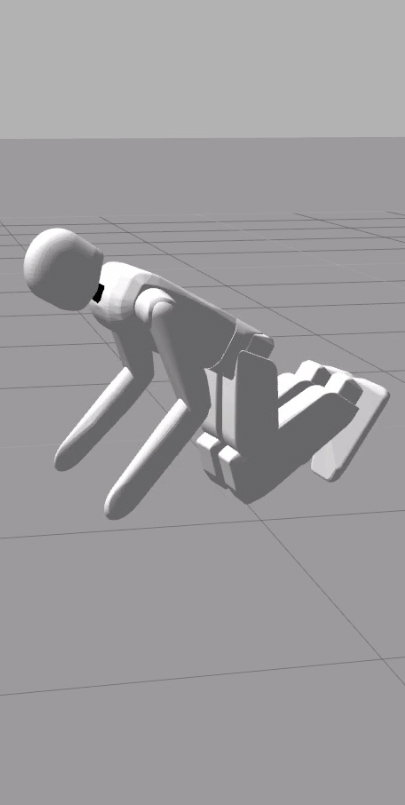}\hspace{0.5px}
\includegraphics[height=0.295\linewidth]{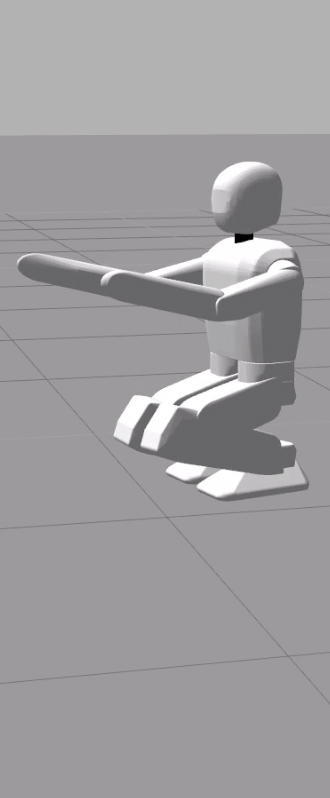}\hspace{0.5px}
\includegraphics[height=0.295\linewidth]{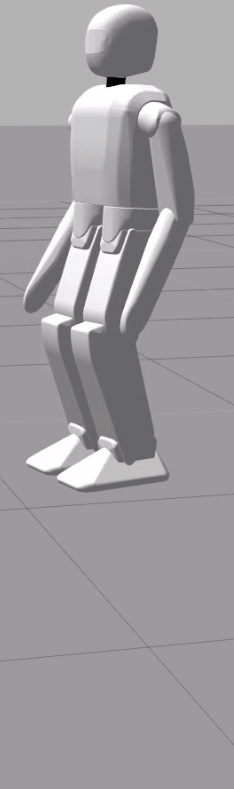}\\\vspace*{1ex}
\includegraphics[height=0.443\linewidth]{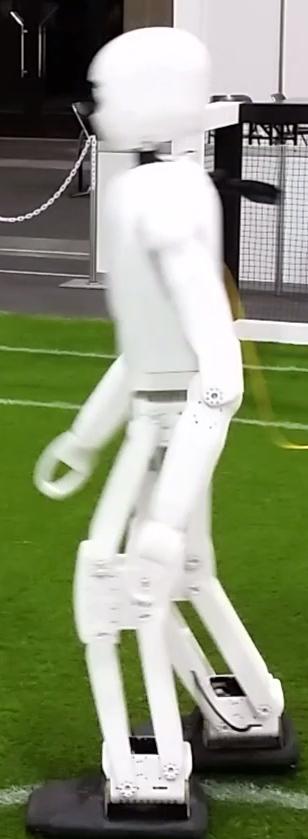}\hspace{1px}
\includegraphics[height=0.443\linewidth]{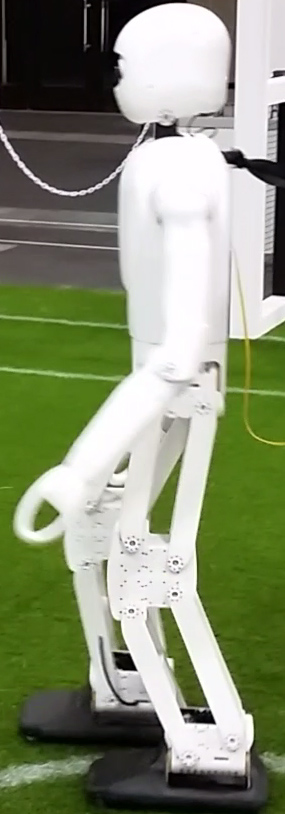}\hspace{1px}
\includegraphics[height=0.443\linewidth]{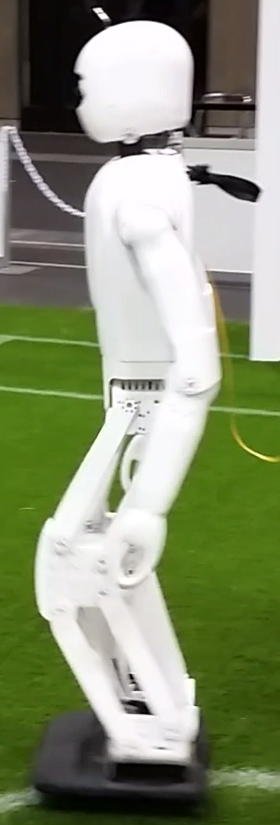}\hspace{1px}
\includegraphics[height=0.443\linewidth]{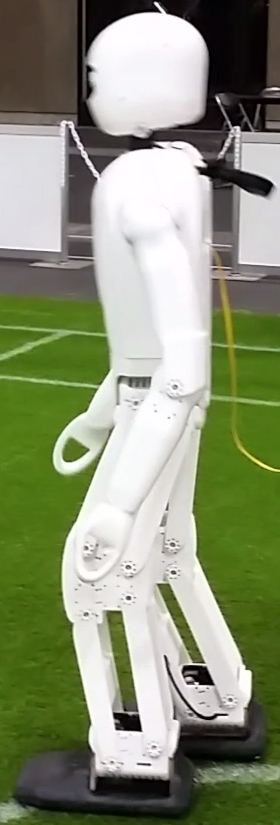}\hspace{1px}
\includegraphics[height=0.443\linewidth]{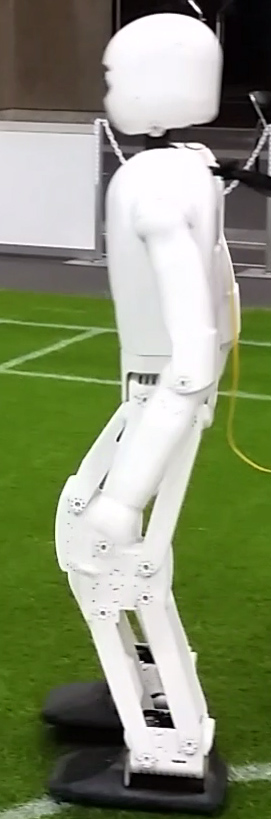}\hspace{1px}
\includegraphics[height=0.443\linewidth]{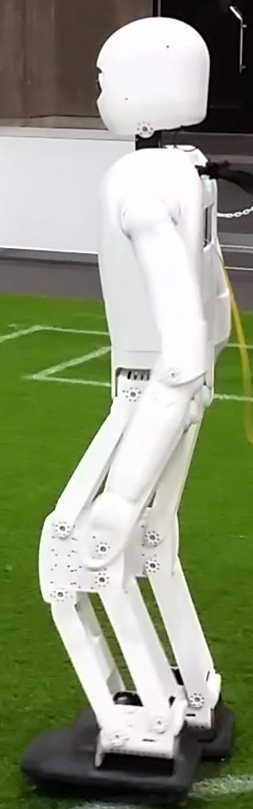}}
\caption{Still frames from captured motions. Top: Get-up motion from prone position in simulation. Bottom: Balanced walking.}
\figlabel{OP2_motions}
\end{figure}

\subsection{Motions}

For playing pre-designed motions, we use the \emph{motion player} which is a part of our software release.
The motion player is a non-linear keyframe interpolator, which creates smooth trajectories between a list of keyframes for joint positions, velocities and efforts. It also supports PID control for specific joints based on the estimated orientation feedback. Creating and editing motions is the role of the \emph{trajectory editor} seen in Fig.~\ref{trajectory_editor}. 
The user-friendly interface allows for intuitive designing of complex motions consisting of multiple frames. 

\begin{figure}[t]
\parbox{\linewidth}{\centering
\includegraphics[width=0.98\linewidth]{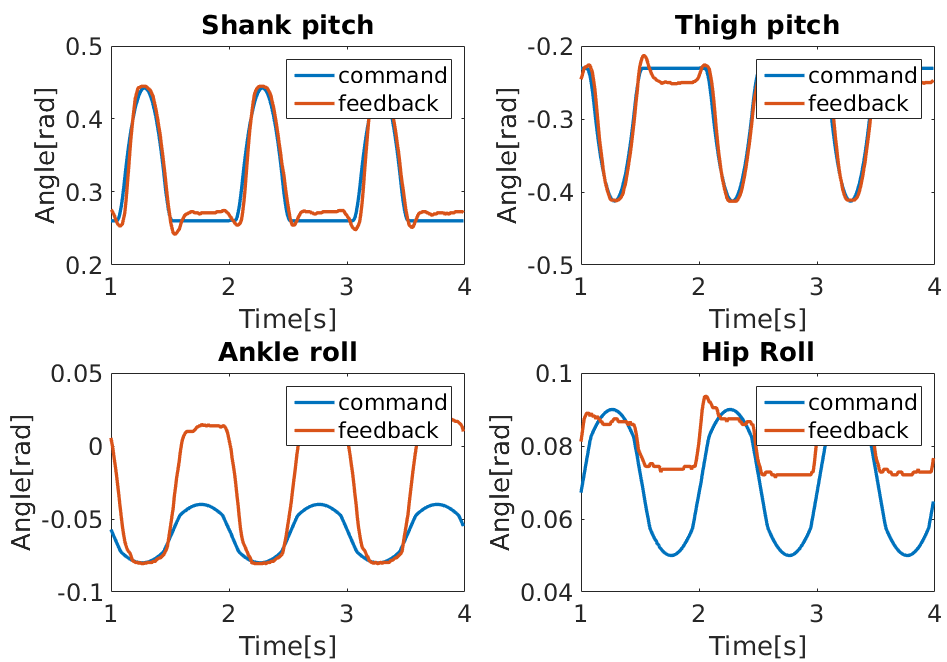}
\includegraphics[width=0.98\linewidth]{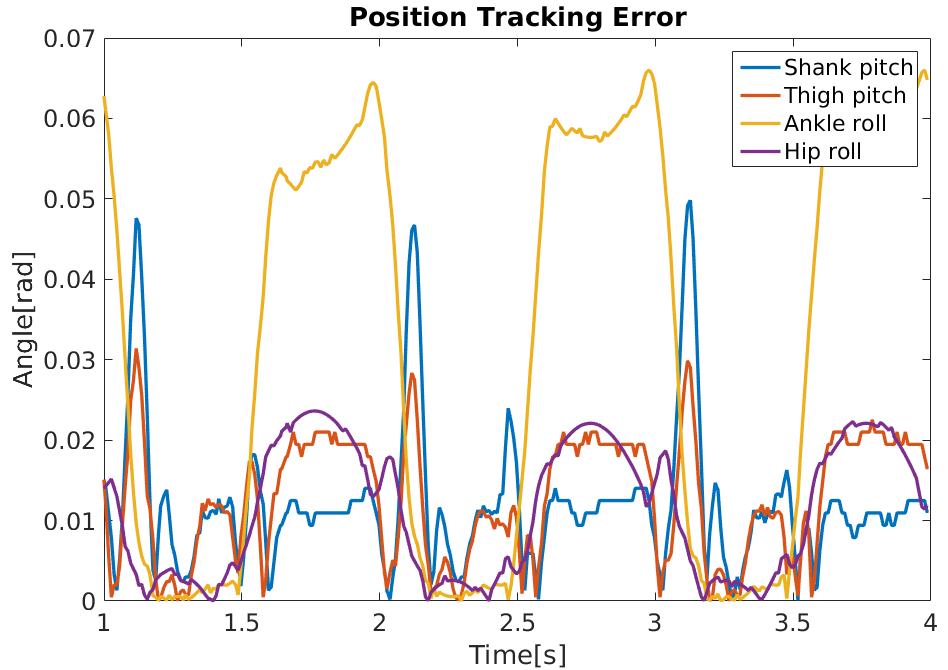}}
\caption{Tracking performance for significantly loaded joints during walking. Top: Trajectory sent and executed by selected joints, Bottom: Tracking error of these joints.}
\figlabel{OP2_tracking}
\end{figure}

\section{Results}

The robot has been tested, both in a controlled and uncontrolled environment. We have initially tuned and tested the robot's gait achieve stable walking at relatively high speed of approximately \SI{0.4}{m/s}.
The tracking performance of the leg joints during walking are shown in \figref{OP2_tracking}.
Under full load, the maximum tracking error varies depending on the joint. In the ankle roll, the 4 degree error is actually the result of a slight overcompensation of the feed-forward position control. This error does not hinder the robot's balance while performing pre-programmed motions nor when walking.
The error in itself is negligible as the artificial grass on which the robot walks provides a dampening effect. However in extreme cases, the error can be handled by the feedback mechanisms present on the robot. 
Other joints display better tracking abilities (below two degrees), as they are located consecutively closer to the CoM of the robot and the feed-forward position control can estimate torques with higher precision. 
To further increase the tracking performance, effort values could still be increased, although this decreases compliance and therefore the life-span of the servos.
\begin{figure}[tbh]
\centering\includegraphics[width=0.98\linewidth]{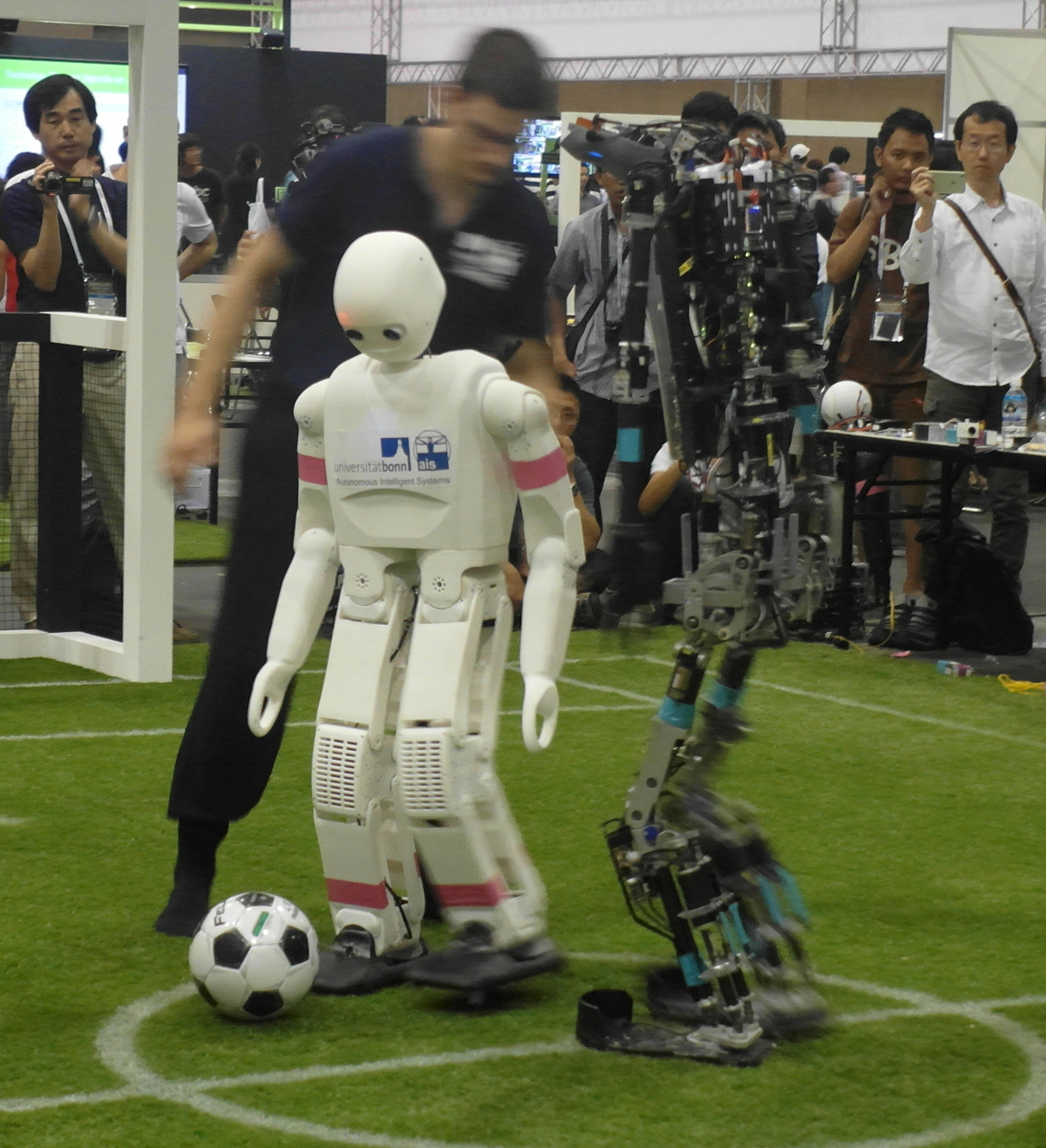}\vspace*{-1ex}
\caption{\nopxl during the final at RoboCup 2017 in Nagoya, Japan.}
\label{robocup}
\end{figure}

We have also performed initial tests in a \emph{gazebo} simulation of the platform, which shows promising results in terms of performing getting up from the supine and prone positions, as well as jumping motions. 
The robot's walk and get-up motion in the simulator are shown in \figref{OP2_motions}.

\begin{figure}[tbh]
\includegraphics[width=\linewidth]{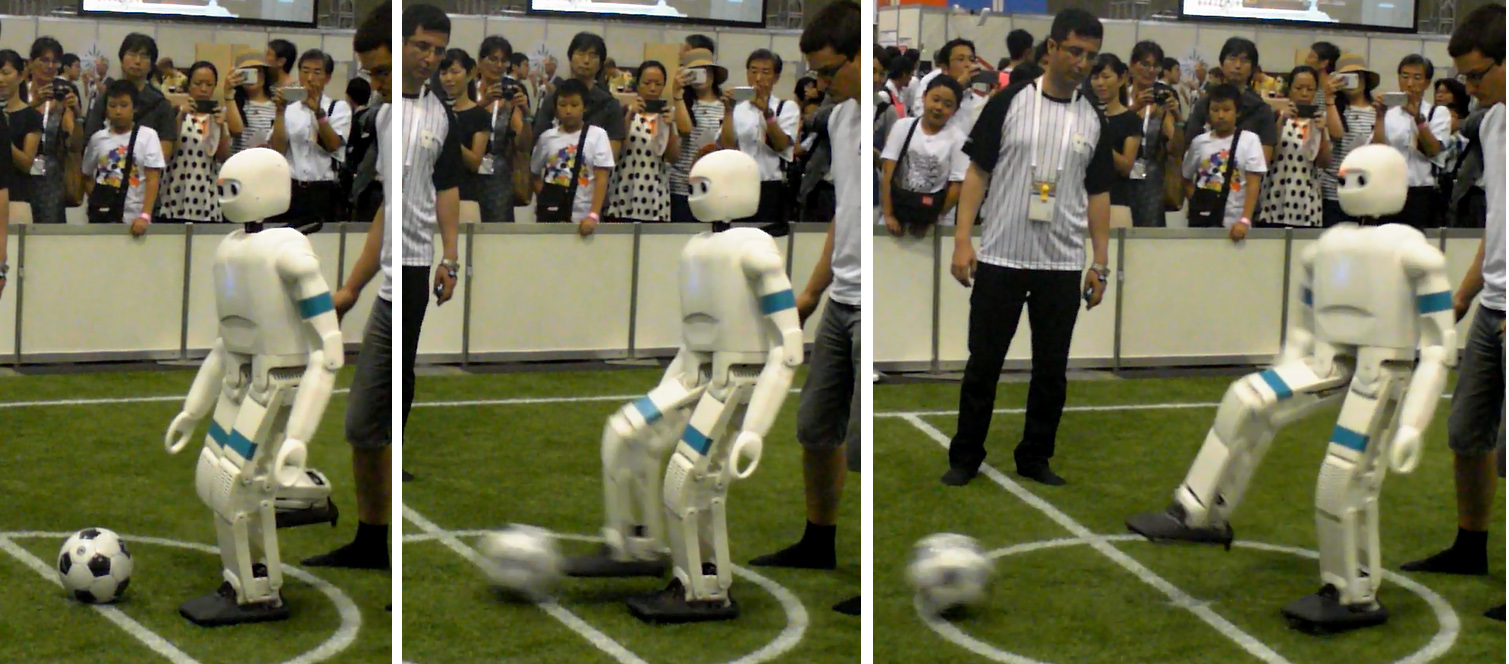}\vspace*{-1ex}
\caption{NimbRo-OP2 performing a kick at RoboCup 2017.}
\label{fig:Kick}
\end{figure}

For our participation in RoboCup 2017, we have designed a reliable kicking motion, which allows the \nopxl to kick the ball on a soccer field farther than any other of our robots, with distances exceeding \SI{6}{m}. 
Fig.~\ref{fig:Kick} shows the robot kicking a regular (size 4) soccer ball.

During the Humanoid League AdultSize competition at RoboCup 2017 in Nagoya, Japan, the NimbRo-OP2 robot proved to be robust, efficient and effective by outperforming all of it's competitors (see Fig.~\ref{robocup}). The robot has won both the soccer competition as well as the technical challenges. 
In five games, the robot has scored 46 goals, while conceding only a single goal in the final, ending the game early with a result of 11:1. 

\begin{figure}[h]
\includegraphics[width=\linewidth]{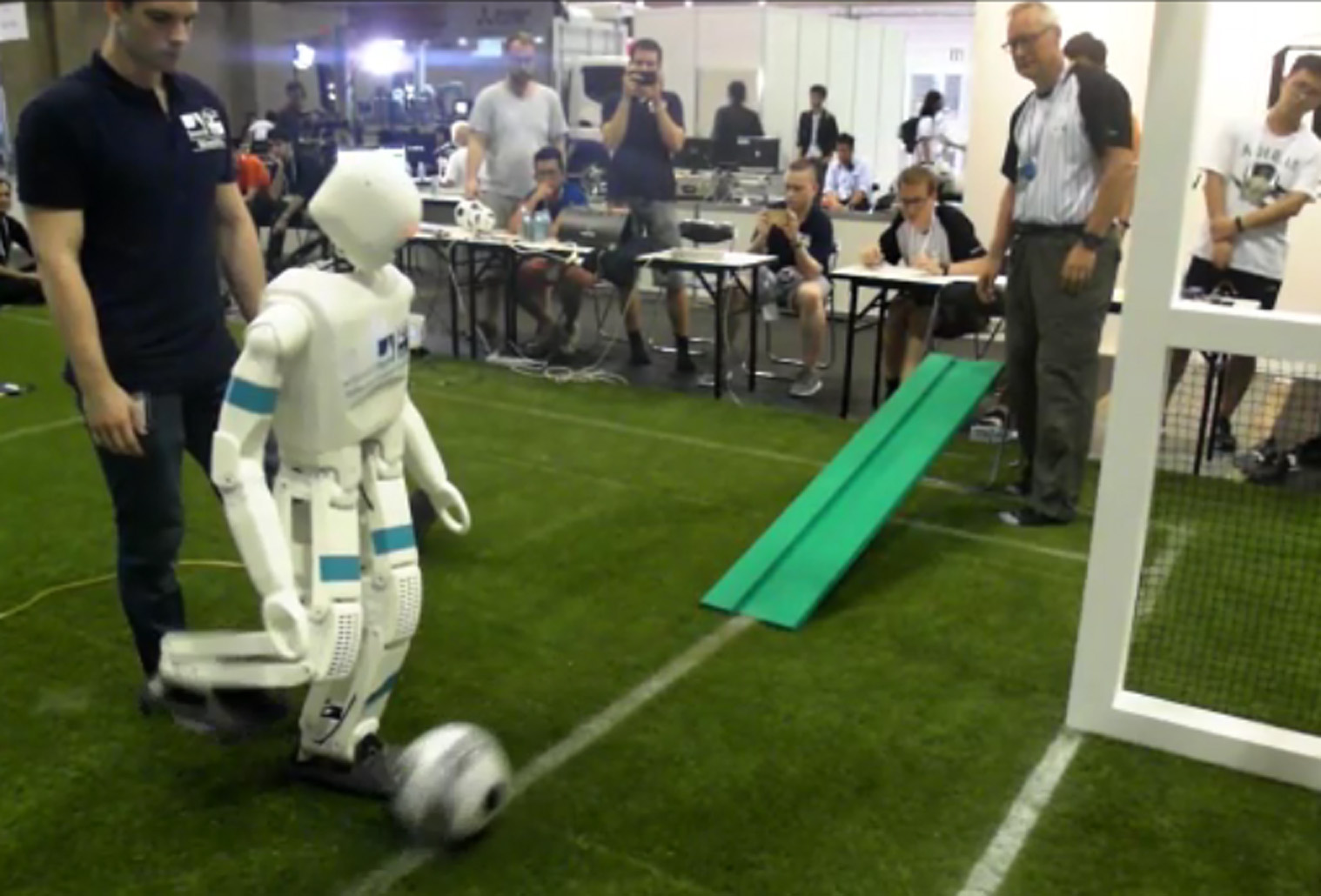}\vspace*{-1ex}
\caption{Kicking a moving ball during RoboCup 2017 Technical Challenge.}
\label{fig:Kick_Moving_Ball}
\end{figure}

\begin{figure}[h]
\includegraphics[width=\linewidth]{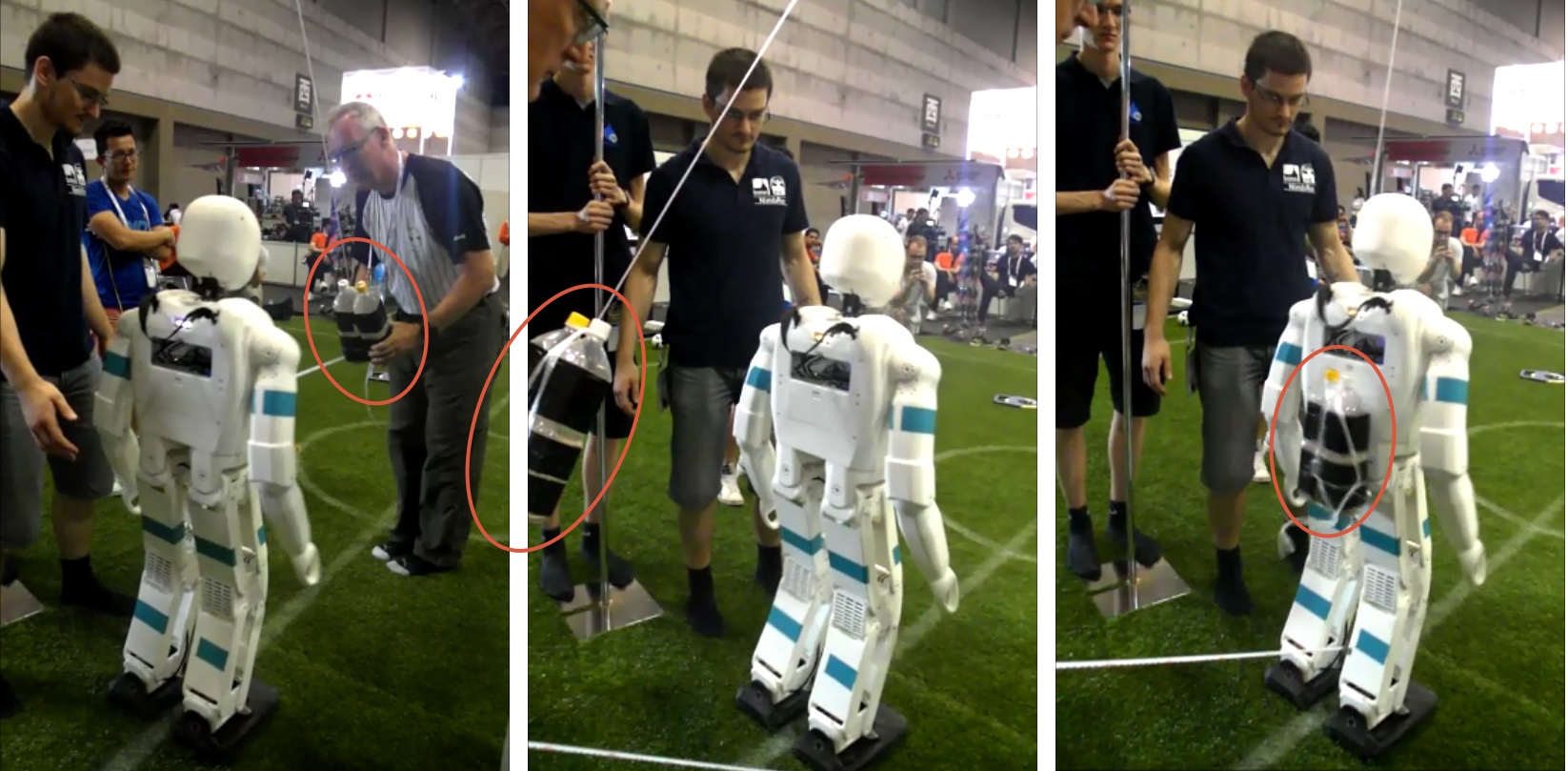}\vspace*{-1ex}
\caption{Resisting impacts during RoboCup 2017 Technical Challenge.}
\label{fig:Impact_Test}
\end{figure}

Our robot won also the Technical Challenges at RoboCup 2017, kicking a moving ball (see Fig.~\ref{fig:Kick_Moving_Ball}) and maintaining walking stability after impacts from a mass on a pendulum (see Fig.~\ref{fig:Impact_Test}). 

The \nopxl has also been awarded the RoboCup Design Award by 
Flower Robotics, based on various criteria which include performance, simplicity, interaction and ease of use. Detailed results can be found on our website\footnote{\url{http://nimbro.net/Humanoid}}.

\section{Conclusions}

The \nopxl is a natural advancement on its predecessors---the \nop and \iguhop. Using the experience we have gained with 3D-printed robots through our cooperation with \igus GmbH on the \iguhop, 
we have designed, built and tested a grown-up open platform that is affordable, easy to operate, with a rich set of features while being comparable in size with robots which are at least one order of magnitude 
more expensive. 

\balance

Building this platform has also proven that our open-source software is modular and applicable to robots of different size and structure. In conjuction with the publication of this paper, we have released 
all the necessary hardware information in the form of printable 3D CAD files~\cite{NOP2Hardware} and updated our already open-source software~\cite{IguhopSoftware} in hope that other research groups benefit and contribute 
to further development of humanoid robotics.

\IEEEtriggercmd{\newpage}
\bibliographystyle{IEEEtran}
\bibliography{IEEEabrv,ms}

\end{document}